\documentclass[11pt,a4paper]{article}
\usepackage[hyperref]{emnlp2020}
\usepackage{times}
\usepackage{latexsym}
\usepackage{color}
\usepackage{tabularx}
\usepackage{booktabs}
\usepackage{multirow}
\usepackage{multicol}
\usepackage{amsfonts}
\usepackage{amsmath}
\usepackage{todonotes}

\usepackage{microtype}

\usepackage{booktabs}

\aclfinalcopy 

\title{Common Sense or World Knowledge? Investigating Adapter-Based Knowledge Injection into Pretrained Transformers}

\author{Anne Lauscher$^\clubsuit$~~Olga Majewska$^\spadesuit$~~Leonardo F. R. Ribeiro$^\diamondsuit$\\\textbf{Iryna Gurevych}$^\diamondsuit$~~\textbf{Nikolai Rozanov}$^\spadesuit$~~\textbf{Goran Glava\v{s}}$^\clubsuit$ \vspace{0.2em}\\
  $^\clubsuit$Data and Web Science Group, University of Mannheim, Germany \\
  $^\spadesuit$Wluper, London, United Kingdom \\
  $^\diamondsuit$Ubiquitous Knowledge Processing (UKP) Lab, TU Darmstadt, Germany \\
  \texttt{\{anne,goran\}@informatik.uni-mannheim.de} \\ 
  \texttt{\{olga,nikolai\}@wluper.com} \\
  \texttt{\href{https://www.ukp.tu-darmstadt.de}{www.ukp.tu-darmstadt.de}}
  }

\date{}

\begin{document}
\maketitle

\begin{abstract}

Following the major success of neural language models (LMs) such as BERT or GPT-2 on a variety of language understanding tasks, recent work focused on injecting (structured) knowledge from external resources into these models. While on the one hand, joint pre-training (i.e., training from scratch, adding objectives based on external knowledge to the primary LM objective) may be prohibitively computationally expensive, post-hoc fine-tuning on external knowledge, on the other hand, may lead to the catastrophic forgetting of distributional knowledge. In this work, we investigate models for complementing the distributional knowledge of BERT with conceptual knowledge from ConceptNet and its corresponding Open Mind Common Sense (OMCS) corpus, respectively, using \textit{adapter training}. While overall results on the GLUE benchmark paint an inconclusive picture, a deeper analysis reveals that our adapter-based models substantially outperform BERT (up to 15-20 performance points) on inference tasks that require the type of conceptual knowledge explicitly present in ConceptNet and OMCS. We also open source all our experiments and relevant code under: \url{https://github.com/wluper/retrograph}.

\end{abstract}

\section{Introduction}

Self-supervised neural models like ELMo \cite{peters2018deep}, BERT \cite{devlin2019bert,liu2019roberta}, GPT \cite{Radford2018ImprovingLU,radford2019language}, or XLNet \cite{yang2019xlnet} have rendered language modeling a very suitable pretraining task for learning language representations that are useful for a wide range of language understanding tasks \cite{wang-etal-2018-glue,wang2019superglue}. 
Although shown versatile w.r.t. the types of knowledge \cite{rogers2020primer} they encode, much like their predecessors -- static word embedding models \cite{mikolov2013distributed,pennington2014glove} -- neural LMs still only ``consume'' the distributional information from large corpora. Yet, a number of structured knowledge sources exist -- knowledge bases (KBs) \cite{suchanek2007yago,auer2007dbpedia} and lexico-semantic networks \cite{miller1995wordnet,liu2004conceptnet,navigli2010babelnet} -- encoding many types of knowledge that are underrepresented in text corpora. 

Starting from this observation, most recent efforts focused on injecting factual \cite{zhang-etal-2019-ernie,liu2019kbert,peters-etal-2019-knowledge} and linguistic knowledge \cite{lauscher2019informing,peters-etal-2019-knowledge} into pretrained LMs and demonstrated the usefulness of such knowledge in language understanding tasks \cite{wang-etal-2018-glue,wang2019superglue}.
\textit{Joint pretraining models}, on the one hand, augment distributional LM objectives with additional objectives based on external resources \cite{Yu:2014,Nguyen:2016acl, lauscher2019informing} and train the extended model from scratch. For models like BERT, this implies computationally expensive retraining from scratch of the encoding transformer network.  
\textit{Post-hoc fine-tuning} models \cite{zhang-etal-2019-ernie,liu2019kbert,peters-etal-2019-knowledge}, on the other hand, use the objectives based on external resources to fine-tune the encoder's parameters, pretrained via distributional LM objectives. If the amount of fine-tuning data is substantial, however, this approach may lead to catastrophic forgetting of distributional knowledge obtained in pretraining \cite{goodfellow2014empirical,kirkpatrick2017overcoming}.     
%

In this work, similar to the concurrent work of \newcite{wang2020kadapters}, we turn to the recently proposed \textit{adapter-based fine-tuning} paradigm \cite{rebuffi-cvpr2018,houlsby2019parameter}, which remedies the shortcomings of both joint pretraining and standard post-hoc fine-tuning. Adapter-based training injects additional parameters into the encoder and only tunes their values: original transformer parameters are kept fixed. Because of this, adapter training preserves the distributional information obtained in LM pretraining, without the need for any distributional (re-)training. 
While \cite{wang2020kadapters} inject factual knowledge from Wikidata \cite{vrandevcic2014wikidata} into BERT, in this work, we investigate two resources that are commonly assumed to contain \textit{general-purpose} and \textit{common sense} knowledge:\footnote{Our results in \S\ref{sec:res} scrutinize this assumption.} ConceptNet \cite{liu2004conceptnet,speer2017conceptnet} and the Open Mind Common Sense (OMCS) corpus \cite{singh2002open}, from which the ConceptNet graph was (semi-)automatically extracted. 
For our first model, dubbed \textsc{CN-Adapt}, we first create a synthetic corpus by randomly traversing the ConceptNet graph and then learn adapter parameters with masked language modelling (MLM) training \cite{devlin2019bert} on that synthetic corpus. For our second model, named \textsc{OM-Adapt}, we learn the adapter parameters via MLM training directly on the OMCS corpus. 

We evaluate both models on the GLUE benchmark, where we observe limited improvements over BERT on a subset of GLUE tasks. However, a more detailed inspection reveals large improvements over the base BERT model (up to 20 Matthews correlation points) on language inference (NLI) subsets labeled as requiring World Knowledge or knowledge about Named Entities. Investigating further, we relate this result to the fact that ConceptNet and OMCS contain much more of what in downstream is considered to be factual world knowledge than what is judged as common sense knowledge. 
Our findings pinpoint the need for more detailed analyses of compatibility between (1) the types of knowledge contained by external resources; and (2) the types of knowledge that benefit concrete downstream tasks; within the emerging body of work on injecting knowledge into pretrained transformers.     

\section{Knowledge Injection Models}

In this work, we are primarily set to investigate if injecting specific types of knowledge (given in the external resource) benefits downstream inference that clearly requires those exact types of knowledge. Because of this, we use the arguably most straightforward mechanisms for injecting the ConceptNet and OMCS information into BERT, and leave the exploration of potentially more effective knowledge injection objectives for future work. 
We inject the external information into adapter parameters of the adapter-augmented BERT \cite{houlsby2019parameter} via BERT's natural objective -- masked language modelling (MLM). OMCS, already a corpus in natural language, is directly subjectable to MLM training -- we filtered out non-English sentences. To subject ConceptNet to MLM training, we need to transform it into a synthetic corpus.  
%


\paragraph{Unwrapping ConceptNet.}

Following established previous work \citep{10.1145/2623330.2623732, ristoski2016rdf2vec}, we induce a synthetic corpus from ConceptNet by randomly traversing its graph. 
We convert relation strings into NL phrases (e.g., \texttt{synonyms} to \emph{is a synonym of}) and duplicate the object node of a triple, using it as the subject for the next sentence. For example, from the path ``\textit{alcoholism}$\xrightarrow{\mathit{causes}}$ \textit{stigma} $\xrightarrow{\textit{hasContext}}$ \textit{christianity} $\xrightarrow{\textit{partOf}}$ \textit{religion}''
we create the text ``\textit{alcoholism causes stigma. stigma is used in the context of christianity. christianity is part of religion.}''. We set the walk lengths to $30$ relations 
%
%
and sample the starting and neighboring nodes from uniform distributions. In total, we performed 2,268,485 walks, resulting with the corpus of 34,560,307
synthetic sentences.

\normalsize






\paragraph{Adapter-Based Training.}
We follow \citet{houlsby2019parameter} and adopt the adapter-based architecture for which they report solid performance across the board. We inject \textit{bottleneck adapters} into BERT's transformer layers. In each transformer layer, we insert two bottleneck adapters: one after the multi-head attention sub-layer and another after the feed-forward sub-layer. Let $\mathbf{X} \in \mathbb{R}^{T \times H}$ be the sequence of contextualized vectors (of size $H$) for the input of $T$ tokens in some transformer layer, input to a bottleneck adapter. The bottleneck adapter, consisting of two feed-forward layers and a residual connection, yields the following output: 
\begin{equation*}
    \mathit{Adapter}(\mathbf{X}) = \mathbf{X} + f\left(\mathbf{X} \mathbf{W}_d + \mathbf{b}_d\right)\mathbf{W}_u + \textbf{b}_u
\end{equation*}
\noindent where $\mathbf{W}_d$ (with bias $\mathbf{b}_d$) and $\mathbf{W}_u$ (with bias $\mathbf{b}_u$) are adapter's parameters, that is, the weights of the linear down-projection and up-projection sub-layers and $f$ is the non-linear activation function. Matrix $\mathbf{W}_d \in \mathbb{R}^{H \times m}$ compresses vectors in $\mathbf{X}$ to the \textit{adapter size} $m < H$, and the matrix $\mathbf{W}_u \in \mathbb{R}^{m \times H}$ projects the activated down-projections back to transformer's hidden size $H$. 
The ratio $H/m$ determines how many times fewer parameters we optimize with adapter-based training compared to standard fine-tuning of all transformer's parameters. 



\section{Evaluation}

We first briefly describe the downstream tasks and training details, and then proceed with the discussion of results obtained with our adapter models. 

\subsection{Experimental Setup.} 

\paragraph{Downstream Tasks.}

We evaluate BERT and our two adapter-based models, \textsc{CN-Adapt} and \textsc{OM-Adapt}, with injected knowledge from ConceptNet and OMCS, respectively, on the tasks from the GLUE benchmark \citep{wang-etal-2018-glue}: 

\vspace{1.4mm}

\noindent \textbf{CoLA} \cite{warstadt2018neural}: Binary sentence classification, predicting grammatical acceptability of sentences from linguistic publications;


\vspace{1.3mm}
\noindent \textbf{SST-2} \cite{socher2013recursive}: Binary sentence classification, predicting binary sentiment (positive or negative) for movie review sentences;

\vspace{1.3mm}
\noindent \textbf{MRPC} \cite{dolan2005automatically}: Binary sentence-pair classification, recognizing sentences which are are mutual paraphrases; 

\vspace{1.3mm}
\noindent \textbf{STS-B} \cite{cer-etal-2017-semeval}: Sentence-pair regression task, predicting the degree of semantic similarity for a given pair of sentences;


\vspace{1.3mm}
\noindent \textbf{QQP} \cite{chen2018quora}: Binary classification task, recognizing question paraphrases;

\vspace{1.3mm}
\noindent \textbf{MNLI} \cite{williams2018broad}: Ternary natural language inference (NLI) classification of sentence pairs. Two test sets are given: a matched version (MNLI-m) in which the test domains match the domains from training data, and a mismatched version (MNLI-mm) with different test domains;

\vspace{1.3mm}
\noindent \textbf{QNLI}: A binary classification version of the Stanford Q\&A dataset \citep{rajpurkar-etal-2016-squad}; 

\vspace{1.3mm}
\noindent \textbf{RTE} \cite{bentivogli2009fifth}: Another NLI dataset, ternary entailment classification for sentence pairs; 

\vspace{1.3mm}
\noindent \textbf{Diag} \cite{wang-etal-2018-glue}: A manually curated NLI dataset, with examples labeled with specific types of knowledge needed for entailment decisions.

\paragraph{Training Details.}
We inject our adapters into a BERT Base model ($12$ transformer layers with $12$ attention heads each; $H = 768$) pretrained on lowercased corpora. Following \cite{houlsby2019parameter}, we set the size of all adapters to $m = 64$ and use GELU  \cite{hendrycks2016gaussian} as the adapter activation $f$. We train the adapter parameters with the Adam algorithm \cite{kingma2015adam} (initial learning rate set to $1e^{-4}$, with $10000$ warm-up steps and the weight decay factor of $0.01$). In downstream fine-tuning, we train in batches of size $16$ and limit the input sequences to $T = 128$ wordpiece tokens. For each task, we find the optimal hyperparameter configuration from the following grid: learning rate $l \in \{2\cdot 10^{-5}, 3\cdot10^{-5}\}$, epochs in $n \in \{3, 4\}$. 
%

%
%


\subsection{Results and Analysis}
\label{sec:res}

\paragraph{GLUE Results.}

\setlength{\tabcolsep}{5pt}
\begin{table*}[t]
\centering
\small{
\begin{tabularx}{\linewidth}{l c c c c c c c c c c | c}
\toprule
Model & CoLA & SST-2 & MRPC & STS-B & QQP & MNLI-m & MNLI-mm & QNLI & RTE & Diag & Avg\\
   & MCC & Acc & F1 & Spear & F1 & Acc & Acc & Acc & Acc & MCC & --\\
    \midrule
BERT Base & 52.1 & 93.5 & \textbf{88.9} & 85.8 & 71.2 & \textbf{84.6} & 83.4 & 90.5 & 66.4 & 34.2 & 75.1 \\
\midrule
\textsc{OM-Adapt} (25K) & 49.5 & 93.5 & 88.8 & 85.1 & 71.4 & 84.4 & 83.5 & \textbf{90.9} &  67.5 & 35.7 & 75.0 \\
\textsc{OM-Adapt} (100K) & \textbf{53.5} & 93.4 & 87.9 & \textbf{85.9} & 71.1 & 84.2 & \textbf{83.7} & 90.6 & 68.2 & 34.8 & 75.3\\
\midrule
\textsc{CN-Adapt} (50K)  & 49.8 & \textbf{93.9} & \textbf{88.9} & 85.8 & \textbf{71.6} & 84.2 & 83.3 & 90.6 & \textbf{69.7} & 37.0 & \textbf{75.5} \\
\textsc{CN-Adapt} (100K) & 48.8 & 92.8 & 87.1 & 85.7 & 71.5 & 83.9 & 83.2 & 90.8 & 64.1 & \textbf{37.8} & 74.6\\
\bottomrule
\end{tabularx}
}
\caption{Results on test portions of GLUE benchmark tasks. Numbers in brackets next to adapter-based models (25K, 50K, 100K) indicate the number of update steps of adapter training on the synthetic ConceptNet corpus (for \textsc{CN-Adapt}) or on the original OMCS corpus (for \textsc{OM-Adapt}). \textbf{Bold}: the best score in each column.}
\label{tbl:results}
\end{table*}

Table \ref{tbl:results} reveals the performance of \textsc{CN-Adapt} and \textsc{OM-Adapt} in comparison with BERT Base on GLUE evaluation tasks. We show the results for two snapshots of \textsc{OM-Adapt}, after 25K and 100K update steps, and for two snapshots of \textsc{CN-Adapt}, after 50K and 100K steps of adapter training.   
Overall, none of our adapter-based models with injected external knowledge from ConceptNet or OMCS yields significant improvements over BERT Base on GLUE.
%
However, we observe substantial improvements (of around 3 points) on RTE and on the Diagnostics NLI dataset (Diag), which encompasses inference instances that require a specific type of knowledge. 

Since our adapter models draw specifically on the conceptual knowledge encoded in ConceptNet and OMCS, we expect the positive impact of injected external knowledge -- assuming effective injection -- to be most observable on test instances that target the same types of conceptual knowledge. To investigate this further, we measure the model performance across different categories of the Diagnostic NLI dataset. This allows us to tease apart inference instances which truly test the efficacy of our knowledge injection methods. We show the results obtained on different categories of the Diagnostic NLI dataset in Table \ref{tbl:diagnostics-results}. 
\setlength{\tabcolsep}{5pt}
\begin{table}[t]
\centering
\small{
\begin{tabularx}{\linewidth}{l c c c c | c}
\toprule
Model & LS & KNO & LOG & PAS & All \\ \midrule
BERT Base & 38.5 & 20.2 & 26.7 & 39.6 & 34.2 \\ \midrule
\textsc{OM-Adapt} (25K) & 39.1 & \textbf{27.1} & 26.1 & 39.5 & 35.7 \\
\textsc{OM-Adapt} (100K) & 37.5 & 21.2 & 27.4 & 41.0 & 34.8 \\ \midrule
\textsc{CN-Adapt} (50K) & 40.2 & 24.3 & 30.1 & \textbf{42.7} & 37.0  \\
\textsc{CN-Adapt} (100K) & \textbf{44.2} & 25.2 & \textbf{30.4} & 41.9 & 37.8 \\ \bottomrule
\end{tabularx}
}
\caption{Breakdown of Diagnostics NLI performance (Matthews correlation), according to information type needed for inference (coarse-grained categories): Lexical Semantics (LS), Knowledge (KNO), Logic (LOG), and Predicate Argument Structure (PAS).}
\label{tbl:diagnostics-results}
\end{table}
The improvements of our adapter-based models over BERT Base on these phenomenon-specific subsections of the Diagnostics NLI dataset are generally much more pronounced: e.g., \textsc{OM-Adapt} (25K) yields a 7\% improvement on inference that requires factual or common sense knowledge (KNO), whereas \textsc{CN-Adapt} (100K) yields a 6\% boost for inference that depends on lexico-semantic knowledge (LS). These results suggest that (1) ConceptNet and OMCS do contain the specific types of knowledge required for these inference categories and that (2) we managed to inject that knowledge into BERT by training adapters on these resources.   


\paragraph{Fine-Grained Knowledge Type Analysis.} In our final analysis, we ``zoom in'' our models' performances on three fine-grained categories of the Diagnostics NLI dataset -- inference instances that require Common Sense Knowledge (CS), World Knowledge (World), and knowledge about Named Entities (NE), respectively. The results for these fine-grained categories are given in Table \ref{tbl:knowledge}.
%
%
%
\setlength{\tabcolsep}{10pt}
\begin{table}[t]
\centering
\small{
\begin{tabularx}{\linewidth}{l c c c c | c}
\toprule
Model & CS & World & NE \\ \midrule
BERT Base & \textbf{29.0} & 10.3 & 15.1 \\ \midrule
\textsc{OM-Adapt} (25K) & 28.5 & 25.3 & 31.4 \\
\textsc{OM-Adapt} (100K) & 24.5 & 17.3 & 22.3 \\ \midrule
\textsc{CN-Adapt} (50K) & 25.6 & 21.1 & 26.0  \\
\textsc{CN-Adapt} (100K) & 24.4 & \textbf{25.6} & \textbf{36.5} \\ \bottomrule
\end{tabularx}
}
%
\caption{Results (Matthews correlation) on Common Sense (CS), World Knowledge (World), and Named Entities (NE) categories of the Diagnostic NLI dataset.}
\label{tbl:knowledge}
\end{table}
These results show an interesting pattern: our adapter-based knowledge-injection models massively outperform BERT Base (up to 15 and 21 MCC points, respectively) for NLI instances labeled as requiring World Knowledge or knowledge about Named Entities. In contrast, we see drops in performance on instances labeled as requiring common sense knowledge. This initially came as a surprise, given the common belief that OMCS and ConcepNet contain the so-called \textit{common sense} knowledge. 
Manual scrutiny of the diagnostic test instances from both CS and World categories uncovers a noticeable mismatch between the kind of information that is considered common sense in KBs like ConceptNet and what is considered common sense knowledge in the downstream. In fact, the majority of information present in ConceptNet and OMCS falls under the World Knowledge definition of the Diagnostic NLI dataset, including factual geographic information (\texttt{stockholm [partOf] sweden}), domain knowledge (\texttt{roadster [isA] car}) and specialized terminology (\texttt{indigenous [synonymOf]	aboriginal}).
%
%
\begin{table*}[t]
    \centering
    \def\arraystretch{1.0}
    \renewcommand\tabcolsep{3pt}
    \begin{tabularx}{1.0\textwidth}{lXXX}
    \toprule
        {Knowledge} & {Premise} & {Hypothesis} & {ConceptNet?} \\
        \midrule
        {World} & {\textit{The sides came to an agreement after their meeting in \textbf{Stockholm}.}} & {\textit{The sides came to an agreement after their meeting in \textbf{Sweden}.}} & {\texttt{\footnotesize stockholm [partOf] sweden}} \\
        {} & {\textit{Musk decided to offer up his personal Tesla \textbf{roadster}.}} & {\textit{Musk decided to offer up his personal \textbf{car}.}} & {\texttt{\footnotesize roadster [isA] car}} \\
        {} & {\textit{The Sydney area has been inhabited by \textbf{indigenous} Australians for at least 30,000 years.}} & {\textit{The Sydney area has been inhabited by \textbf{Aboriginal} people for at least 30,000 years.}} & {\texttt{\footnotesize indigenous [synonymOf]	aboriginal}} \\
        \midrule
        {Common Sense} & {\textit{My jokes fully reveal my character.}} & {\textit{If everyone believed my jokes, they'd know exactly who I was.}} & {}  \\
        {} & {\textit{The systems thus produced are incremental: dialogues are processed word-by-word, shown previously to be essential in supporting natural, spontaneous dialogue.}} & {\textit{The systems thus produced support the capability to interrupt an interlocutor mid-sentence.}} & {}  \\
        {} & {\textit{He deceitfully proclaimed: ``This is all I ever really wanted."}} & {\textit{He was satisfied.}} & {}  \\
        \bottomrule
    
    \end{tabularx}%
    \caption{Premise-hypothesis examples from the diagnostic NLI dataset tagged for commonsense and world knowledge, and relevant ConceptNet relations, where available.} 
    \label{tbl:examples}
\end{table*}
In contrast, many of the CS inference instances require complex, high-level reasoning, understanding metaphorical and idiomatic meaning, and making far-reaching connections. We display NLI Dignostics examples from the World Knowledge and Common Sense categories in Table \ref{tbl:examples}. 
In such cases, explicit conceptual links often do not suffice for a correct inference and much of the required knowledge is not explicitly encoded in the external resources. 
Consider, e.g., the following CS NLI instance:  [\texttt{premise:} \textit{My jokes fully reveal my character} \texttt{; hypothesis:} \textit{If everyone believed   my jokes, they’d know exactly who I was} \texttt{; entailment}]. While ConceptNet and OMCS may associate \textit{character} with \textit{personality} or \textit{personality} with \textit{identity}, the knowledge that the phrase \textit{who I was} may refer to \textit{identity} is beyond the explicit knowledge present in these resources.
This sheds light on the results in Table \ref{tbl:knowledge}: when the knowledge required to tackle the inference problem at hand is available in the external resource, our adapter-based knowledge-injected models significantly outperform the baseline transformer; otherwise, the benefits of knowledge injection are negligible or non-existent. The promising results on \textit{world knowledge} and \textit{named entities} portions of the Diagnostics dataset suggest that our methods does successfully inject external information into the pretrained transformer and that the presence of the required knowledge for the task in the external resources is an obvious prerequisite.




\section{Conclusion}

We presented two simple strategies for injecting external knowledge from ConceptNet and OMCS corpus, respectively, into BERT via bottleneck adapters. Additional adapter parameters store the external knowledge and allow for the preservation of the rich distributional knowledge acquired in BERT's pretraining in the original transformer parameters. We demonstrated the effectiveness of these models in language understanding settings that require precisely the type of knowledge that one finds in ConceptNet and OMCS, in which our adapter-based models outperform BERT by up to 20 performance points. Our findings stress the importance of having detailed analyses that compare (a) the types of knowledge found in external resources being injected against (b) the types of knowledge that a concrete downstream reasoning tasks requires. We hope this work motivates further research effort in the direction of fine-grained knowledge typing, both of explicit knowledge in external resources and the implicit knowledge stored in pretrained transformers.

\section*{Acknowledgments}

Anne Lauscher and Goran Glavaš are supported by the Eliteprogramm of the Baden-Württemberg Stiftung (AGREE grant). Leonardo F. R. Ribeiro has been supported by the German Research Foundation as part of the Research Training Group AIPHES under the grant No. GRK 1994/1. This work has been supported by the German Research Foundation within the project “Open Argument Mining” (GU 798/25-1), associated with the Priority Program “Robust Argumentation Machines (RATIO)” (SPP-1999). The work of Olga Majewska was conducted under the research lab of Wluper Ltd. (UK/ 10195181).

\bibliography{references}

\begin{thebibliography}{42}
\expandafter\ifx\csname natexlab\endcsname\relax\def\natexlab#1{#1}\fi

\bibitem[{Auer et~al.(2007)Auer, Bizer, Kobilarov, Lehmann, Cyganiak, and
  Ives}]{auer2007dbpedia}
S{\"o}ren Auer, Christian Bizer, Georgi Kobilarov, Jens Lehmann, Richard
  Cyganiak, and Zachary Ives. 2007.
\newblock Dbpedia: A nucleus for a web of open data.
\newblock In \emph{The semantic web}, pages 722--735. Springer.

\bibitem[{Bentivogli et~al.(2009)Bentivogli, Clark, Dagan, and
  Giampiccolo}]{bentivogli2009fifth}
Luisa Bentivogli, Peter Clark, Ido Dagan, and Danilo Giampiccolo. 2009.
\newblock The fifth pascal recognizing textual entailment challenge.
\newblock In \emph{TAC}.

\bibitem[{Cer et~al.(2017)Cer, Diab, Agirre, Lopez-Gazpio, and
  Specia}]{cer-etal-2017-semeval}
Daniel Cer, Mona Diab, Eneko Agirre, I{\~n}igo Lopez-Gazpio, and Lucia Specia.
  2017.
\newblock \href {https://doi.org/10.18653/v1/S17-2001} {{S}em{E}val-2017 task
  1: Semantic textual similarity multilingual and crosslingual focused
  evaluation}.
\newblock In \emph{Proceedings of the 11th International Workshop on Semantic
  Evaluation ({S}em{E}val-2017)}, pages 1--14, Vancouver, Canada. Association
  for Computational Linguistics.

\bibitem[{Chen et~al.(2018)Chen, Zhang, Zhang, and Zhao}]{chen2018quora}
Zihan Chen, Hongbo Zhang, Xiaoji Zhang, and Leqi Zhao. 2018.
\newblock Quora question pairs.

\bibitem[{Devlin et~al.(2019)Devlin, Chang, Lee, and
  Toutanova}]{devlin2019bert}
Jacob Devlin, Ming-Wei Chang, Kenton Lee, and Kristina Toutanova. 2019.
\newblock Bert: Pre-training of deep bidirectional transformers for language
  understanding.
\newblock In \emph{Proceedings of the 2019 Conference of the North American
  Chapter of the Association for Computational Linguistics: Human Language
  Technologies, Volume 1 (Long and Short Papers)}, pages 4171--4186.

\bibitem[{Dolan and Brockett(2005)}]{dolan2005automatically}
William~B Dolan and Chris Brockett. 2005.
\newblock Automatically constructing a corpus of sentential paraphrases.
\newblock In \emph{Proceedings of the Third International Workshop on
  Paraphrasing (IWP2005)}.

\bibitem[{Goodfellow et~al.(2014)Goodfellow, Mirza, Da~Xiao, and
  Bengio}]{goodfellow2014empirical}
Ian~J Goodfellow, Mehdi Mirza, Aaron~Courville Da~Xiao, and Yoshua Bengio.
  2014.
\newblock An empirical investigation of catastrophic forgeting in gradientbased
  neural networks.
\newblock In \emph{In Proceedings of International Conference on Learning
  Representations (ICLR}. Citeseer.

\bibitem[{Hendrycks and Gimpel(2016)}]{hendrycks2016gaussian}
Dan Hendrycks and Kevin Gimpel. 2016.
\newblock \href {http://arxiv.org/abs/1606.08415} {Gaussian error linear units
  (gelus)}.

\bibitem[{Houlsby et~al.(2019)Houlsby, Giurgiu, Jastrzebski, Morrone,
  De~Laroussilhe, Gesmundo, Attariyan, and Gelly}]{houlsby2019parameter}
Neil Houlsby, Andrei Giurgiu, Stanislaw Jastrzebski, Bruna Morrone, Quentin
  De~Laroussilhe, Andrea Gesmundo, Mona Attariyan, and Sylvain Gelly. 2019.
\newblock Parameter-efficient transfer learning for nlp.
\newblock In \emph{International Conference on Machine Learning}, pages
  2790--2799.

\bibitem[{Kingma and Ba(2015)}]{kingma2015adam}
Diederik~P Kingma and Jimmy Ba. 2015.
\newblock Adam: A method for stochastic optimization.
\newblock In \emph{Proceedings of ICLR}.

\bibitem[{Kirkpatrick et~al.(2017)Kirkpatrick, Pascanu, Rabinowitz, Veness,
  Desjardins, Rusu, Milan, Quan, Ramalho, Grabska-Barwinska
  et~al.}]{kirkpatrick2017overcoming}
James Kirkpatrick, Razvan Pascanu, Neil Rabinowitz, Joel Veness, Guillaume
  Desjardins, Andrei~A Rusu, Kieran Milan, John Quan, Tiago Ramalho, Agnieszka
  Grabska-Barwinska, et~al. 2017.
\newblock Overcoming catastrophic forgetting in neural networks.
\newblock \emph{Proceedings of the national academy of sciences},
  114(13):3521--3526.

\bibitem[{Lauscher et~al.(2019)Lauscher, Vuli{\'c}, Ponti, Korhonen, and
  Glava{\v{s}}}]{lauscher2019informing}
Anne Lauscher, Ivan Vuli{\'c}, Edoardo~Maria Ponti, Anna Korhonen, and Goran
  Glava{\v{s}}. 2019.
\newblock Informing unsupervised pretraining with external linguistic
  knowledge.
\newblock \emph{arXiv preprint arXiv:1909.02339}.

\bibitem[{Liu and Singh(2004)}]{liu2004conceptnet}
Hugo Liu and Push Singh. 2004.
\newblock Conceptnet—a practical commonsense reasoning tool-kit.
\newblock \emph{BT technology journal}, 22(4):211--226.

\bibitem[{Liu et~al.(2019{\natexlab{a}})Liu, Zhou, Zhao, Wang, Ju, Deng, and
  Wang}]{liu2019kbert}
Weijie Liu, Peng Zhou, Zhe Zhao, Zhiruo Wang, Qi~Ju, Haotang Deng, and Ping
  Wang. 2019{\natexlab{a}}.
\newblock K-bert: Enabling language representation with knowledge graph.
\newblock \emph{arXiv preprint arXiv:1909.07606}.

\bibitem[{Liu et~al.(2019{\natexlab{b}})Liu, Ott, Goyal, Du, Joshi, Chen, Levy,
  Lewis, Zettlemoyer, and Stoyanov}]{liu2019roberta}
Yinhan Liu, Myle Ott, Naman Goyal, Jingfei Du, Mandar Joshi, Danqi Chen, Omer
  Levy, Mike Lewis, Luke Zettlemoyer, and Veselin Stoyanov. 2019{\natexlab{b}}.
\newblock Ro{BERT}a: A robustly optimized bert pretraining approach.
\newblock \emph{arXiv preprint arXiv:1907.11692}.

\bibitem[{Mikolov et~al.(2013)Mikolov, Sutskever, Chen, Corrado, and
  Dean}]{mikolov2013distributed}
Tomas Mikolov, Ilya Sutskever, Kai Chen, Greg~S Corrado, and Jeff Dean. 2013.
\newblock Distributed representations of words and phrases and their
  compositionality.
\newblock In \emph{Advances in neural information processing systems}, pages
  3111--3119.

\bibitem[{Miller(1995)}]{miller1995wordnet}
George~A Miller. 1995.
\newblock Wordnet: a lexical database for english.
\newblock \emph{Communications of the ACM}, 38(11):39--41.

\bibitem[{Navigli and Ponzetto(2010)}]{navigli2010babelnet}
Roberto Navigli and Simone~Paolo Ponzetto. 2010.
\newblock Babelnet: Building a very large multilingual semantic network.
\newblock In \emph{Proceedings of the 48th annual meeting of the association
  for computational linguistics}, pages 216--225. Association for Computational
  Linguistics.

\bibitem[{Nguyen et~al.(2016)Nguyen, Schulte~im Walde, and Vu}]{Nguyen:2016acl}
Kim~Anh Nguyen, Sabine Schulte~im Walde, and Ngoc~Thang Vu. 2016.
\newblock \href {http://anthology.aclweb.org/P16-2074} {Integrating
  distributional lexical contrast into word embeddings for antonym-synonym
  distinction}.
\newblock In \emph{Proceedings of ACL}, pages 454--459.

\bibitem[{Pennington et~al.(2014)Pennington, Socher, and
  Manning}]{pennington2014glove}
Jeffrey Pennington, Richard Socher, and Christopher Manning. 2014.
\newblock Glove: Global vectors for word representation.
\newblock In \emph{Proceedings of the 2014 conference on empirical methods in
  natural language processing (EMNLP)}, pages 1532--1543.

\bibitem[{Perozzi et~al.(2014)Perozzi, Al-Rfou, and
  Skiena}]{10.1145/2623330.2623732}
Bryan Perozzi, Rami Al-Rfou, and Steven Skiena. 2014.
\newblock \href {https://doi.org/10.1145/2623330.2623732} {Deepwalk: Online
  learning of social representations}.
\newblock In \emph{Proceedings of the 20th ACM SIGKDD International Conference
  on Knowledge Discovery and Data Mining}, KDD ’14, page 701–710, New York,
  NY, USA. Association for Computing Machinery.

\bibitem[{Peters et~al.(2018)Peters, Neumann, Iyyer, Gardner, Clark, Lee, and
  Zettlemoyer}]{peters2018deep}
Matthew~E Peters, Mark Neumann, Mohit Iyyer, Matt Gardner, Christopher Clark,
  Kenton Lee, and Luke Zettlemoyer. 2018.
\newblock Deep contextualized word representations.
\newblock In \emph{Proceedings of NAACL-HLT}, pages 2227--2237.

\bibitem[{Peters et~al.(2019)Peters, Neumann, Logan, Schwartz, Joshi, Singh,
  and Smith}]{peters-etal-2019-knowledge}
Matthew~E. Peters, Mark Neumann, Robert Logan, Roy Schwartz, Vidur Joshi,
  Sameer Singh, and Noah~A. Smith. 2019.
\newblock Knowledge enhanced contextual word representations.
\newblock In \emph{Proceedings of the 2019 Conference on Empirical Methods in
  Natural Language Processing and the 9th International Joint Conference on
  Natural Language Processing (EMNLP-IJCNLP)}, pages 43--54.

\bibitem[{Radford et~al.(2018)Radford, Narasimhan, Salimans, and
  Sutskever}]{Radford2018ImprovingLU}
Alec Radford, Karthik Narasimhan, Tim Salimans, and Ilya Sutskever. 2018.
\newblock \href
  {https://www.cs.ubc.ca/~amuham01/LING530/papers/radford2018improving.pdf}
  {Improving language understanding by generative pre-training}.
\newblock \emph{OpenAI Technical Report}.

\bibitem[{Radford et~al.(2019)Radford, Wu, Child, Luan, Amodei, and
  Sutskever}]{radford2019language}
Alec Radford, Jeffrey Wu, Rewon Child, David Luan, Dario Amodei, and Ilya
  Sutskever. 2019.
\newblock Language models are unsupervised multitask learners.
\newblock \emph{OpenAI Blog}, 1(8).

\bibitem[{Rajpurkar et~al.(2016)Rajpurkar, Zhang, Lopyrev, and
  Liang}]{rajpurkar-etal-2016-squad}
Pranav Rajpurkar, Jian Zhang, Konstantin Lopyrev, and Percy Liang. 2016.
\newblock \href {https://doi.org/10.18653/v1/D16-1264} {{SQ}u{AD}: 100,000+
  questions for machine comprehension of text}.
\newblock In \emph{Proceedings of the 2016 Conference on Empirical Methods in
  Natural Language Processing}, pages 2383--2392, Austin, Texas. Association
  for Computational Linguistics.

\bibitem[{Rebuffi et~al.(2018)Rebuffi, Bilen, and Vedaldi}]{rebuffi-cvpr2018}
Sylvestre-Alvise Rebuffi, Hakan Bilen, and Andrea Vedaldi. 2018.
\newblock Efficient parametrization of multi-domain deep neural networks.
\newblock In \emph{CVPR}.

\bibitem[{Ristoski and Paulheim(2016)}]{ristoski2016rdf2vec}
Petar Ristoski and Heiko Paulheim. 2016.
\newblock Rdf2vec: Rdf graph embeddings for data mining.
\newblock In \emph{International Semantic Web Conference}, pages 498--514.
  Springer.

\bibitem[{Rogers et~al.(2020)Rogers, Kovaleva, and
  Rumshisky}]{rogers2020primer}
Anna Rogers, Olga Kovaleva, and Anna Rumshisky. 2020.
\newblock A primer in bertology: What we know about how bert works.
\newblock \emph{arXiv preprint arXiv:2002.12327}.

\bibitem[{Singh et~al.(2002)Singh, Lin, Mueller, Lim, Perkins, and
  Zhu}]{singh2002open}
Push Singh, Thomas Lin, Erik~T Mueller, Grace Lim, Travell Perkins, and Wan~Li
  Zhu. 2002.
\newblock Open mind common sense: Knowledge acquisition from the general
  public.
\newblock In \emph{OTM Confederated International Conferences" On the Move to
  Meaningful Internet Systems"}, pages 1223--1237. Springer.

\bibitem[{Socher et~al.(2013)Socher, Perelygin, Wu, Chuang, Manning, Ng, and
  Potts}]{socher2013recursive}
Richard Socher, Alex Perelygin, Jean Wu, Jason Chuang, Christopher~D Manning,
  Andrew Ng, and Christopher Potts. 2013.
\newblock Recursive deep models for semantic compositionality over a sentiment
  treebank.
\newblock In \emph{Proceedings of the 2013 conference on empirical methods in
  natural language processing}, pages 1631--1642.

\bibitem[{Speer et~al.(2017)Speer, Chin, and Havasi}]{speer2017conceptnet}
Robert Speer, Joshua Chin, and Catherine Havasi. 2017.
\newblock Conceptnet 5.5: An open multilingual graph of general knowledge.
\newblock In \emph{Thirty-First AAAI Conference on Artificial Intelligence}.

\bibitem[{Suchanek et~al.(2007)Suchanek, Kasneci, and
  Weikum}]{suchanek2007yago}
Fabian~M Suchanek, Gjergji Kasneci, and Gerhard Weikum. 2007.
\newblock Yago: a core of semantic knowledge.
\newblock In \emph{Proceedings of the 16th international conference on World
  Wide Web}, pages 697--706. ACM.

\bibitem[{Vrande{\v{c}}i{\'c} and Kr{\"o}tzsch(2014)}]{vrandevcic2014wikidata}
Denny Vrande{\v{c}}i{\'c} and Markus Kr{\"o}tzsch. 2014.
\newblock Wikidata: a free collaborative knowledgebase.
\newblock \emph{Communications of the ACM}, 57(10):78--85.

\bibitem[{Wang et~al.(2019)Wang, Pruksachatkun, Nangia, Singh, Michael, Hill,
  Levy, and Bowman}]{wang2019superglue}
Alex Wang, Yada Pruksachatkun, Nikita Nangia, Amanpreet Singh, Julian Michael,
  Felix Hill, Omer Levy, and Samuel Bowman. 2019.
\newblock Superglue: A stickier benchmark for general-purpose language
  understanding systems.
\newblock In \emph{Advances in Neural Information Processing Systems}, pages
  3261--3275.

\bibitem[{Wang et~al.(2018)Wang, Singh, Michael, Hill, Levy, and
  Bowman}]{wang-etal-2018-glue}
Alex Wang, Amanpreet Singh, Julian Michael, Felix Hill, Omer Levy, and Samuel
  Bowman. 2018.
\newblock \href {https://www.aclweb.org/anthology/W18-5446} {{GLUE}: A
  multi-task benchmark and analysis platform for natural language
  understanding}.
\newblock In \emph{Proceedings of the Blacbox NLP Workshop}, pages 353--355.

\bibitem[{Wang et~al.(2020)Wang, Tang, Duan, Wei, Huang, Cao, Jiang, Zhou
  et~al.}]{wang2020kadapters}
Ruize Wang, Duyu Tang, Nan Duan, Zhongyu Wei, Xuanjing Huang, Cuihong Cao,
  Daxin Jiang, Ming Zhou, et~al. 2020.
\newblock K-adapter: Infusing knowledge into pre-trained models with adapters.
\newblock \emph{arXiv preprint arXiv:2002.01808}.

\bibitem[{Warstadt et~al.(2018)Warstadt, Singh, and
  Bowman}]{warstadt2018neural}
Alex Warstadt, Amanpreet Singh, and Samuel~R Bowman. 2018.
\newblock Neural network acceptability judgments.
\newblock \emph{arXiv preprint arXiv:1805.12471}.

\bibitem[{Williams et~al.(2018)Williams, Nangia, and
  Bowman}]{williams2018broad}
Adina Williams, Nikita Nangia, and Samuel Bowman. 2018.
\newblock A broad-coverage challenge corpus for sentence understanding through
  inference.
\newblock In \emph{Proceedings of the 2018 Conference of the North American
  Chapter of the Association for Computational Linguistics: Human Language
  Technologies, Volume 1 (Long Papers)}, pages 1112--1122.

\bibitem[{Yang et~al.(2019)Yang, Dai, Yang, Carbonell, Salakhutdinov, and
  Le}]{yang2019xlnet}
Zhilin Yang, Zihang Dai, Yiming Yang, Jaime Carbonell, Ruslan Salakhutdinov,
  and Quoc~V Le. 2019.
\newblock Xlnet: Generalized autoregressive pretraining for language
  understanding.
\newblock \emph{arXiv preprint arXiv:1906.08237}.

\bibitem[{Yu and Dredze(2014)}]{Yu:2014}
Mo~Yu and Mark Dredze. 2014.
\newblock \href {http://www.aclweb.org/anthology/P14-2089} {Improving lexical
  embeddings with semantic knowledge}.
\newblock In \emph{Proceedings of ACL}, pages 545--550.

\bibitem[{Zhang et~al.(2019)Zhang, Han, Liu, Jiang, Sun, and
  Liu}]{zhang-etal-2019-ernie}
Zhengyan Zhang, Xu~Han, Zhiyuan Liu, Xin Jiang, Maosong Sun, and Qun Liu. 2019.
\newblock {ERNIE}: Enhanced language representation with informative entities.
\newblock In \emph{Proceedings of the 57th Annual Meeting of the Association
  for Computational Linguistics}, pages 1441--1451.

\end{thebibliography}
\bibliographystyle{acl_natbib}


\end{document}